\documentclass[journal]{IEEEtran}
\usepackage{times}  
\usepackage{helvet} 
\usepackage{courier}  
\usepackage[hyphens]{url}  
\usepackage{graphicx} 
\usepackage{graphicx}  
\usepackage[utf8]{inputenc} 
\usepackage{url}            
\usepackage{booktabs}       
\usepackage{amsfonts}       
\usepackage{nicefrac}       
\usepackage{microtype}      

\usepackage{amsmath,amssymb}
\usepackage{graphicx}
\usepackage{multirow}
\usepackage{array}
\usepackage{subfig}

\usepackage{siunitx}

\title{Video Face Super-Resolution with Motion-Adaptive Feedback Cell}
\author{Jingwei Xin$^\dagger$, Nannan Wang$^\ddagger$, Jie Li$^\dagger$, Xinbo Gao$^\dagger$, Zhifeng Li $^\S $ \\
	$^\dagger$ State Key Laboratory of Integrated Services Networks, \\
	School of Electronic Engineering, Xidian University, Xi'an 710071, China \\
	$^\ddagger$ State Key Laboratory of Integrated Services Networks, \\
	School of Telecommunications Engineering, Xidian University, Xi'an 710071, China \\
	$^\S$ Tencent AI Lab, China, \\
}
 
\begin{document}

\maketitle

\begin{abstract}
Video super-resolution (VSR) methods have recently achieved a remarkable success due to the development of deep convolutional neural networks (CNN). Current state-of-the-art CNN methods usually treat the VSR problem as a large number of separate multi-frame super-resolution tasks, at which a batch of low resolution (LR) frames is utilized to generate a single high resolution (HR) frame, and running a slide window to select LR frames over the entire video would obtain a series of HR frames. However, duo to the complex temporal dependency between frames, with the number of LR input frames increase, the performance of the reconstructed HR frames become worse. The reason is in that these methods lack the ability to model complex temporal dependencies and hard to give an accurate motion estimation and compensation for VSR process. Which makes the performance degrade drastically when the motion in frames is complex. In this paper, we propose a Motion-Adaptive Feedback Cell (MAFC), a simple but effective block, which can efficiently capture the motion compensation and feed it back to the network in an adaptive way. Our approach efficiently utilizes the information of the inter-frame motion, the dependence of the network on motion estimation and compensation method can be avoid. In addition, benefiting from the excellent nature of MAFC, the network can achieve better performance in the case of extremely complex motion scenarios. Extensive evaluations and comparisons validate the strengths of our approach, and the experimental results demonstrated that the proposed framework is outperform the state-of-the-art methods.
\end{abstract}

\begin{figure*}[t]
	\captionsetup{belowskip=-10pt}
	\centering
	\scalebox{0.8}{
	\includegraphics[width=1.0\linewidth]{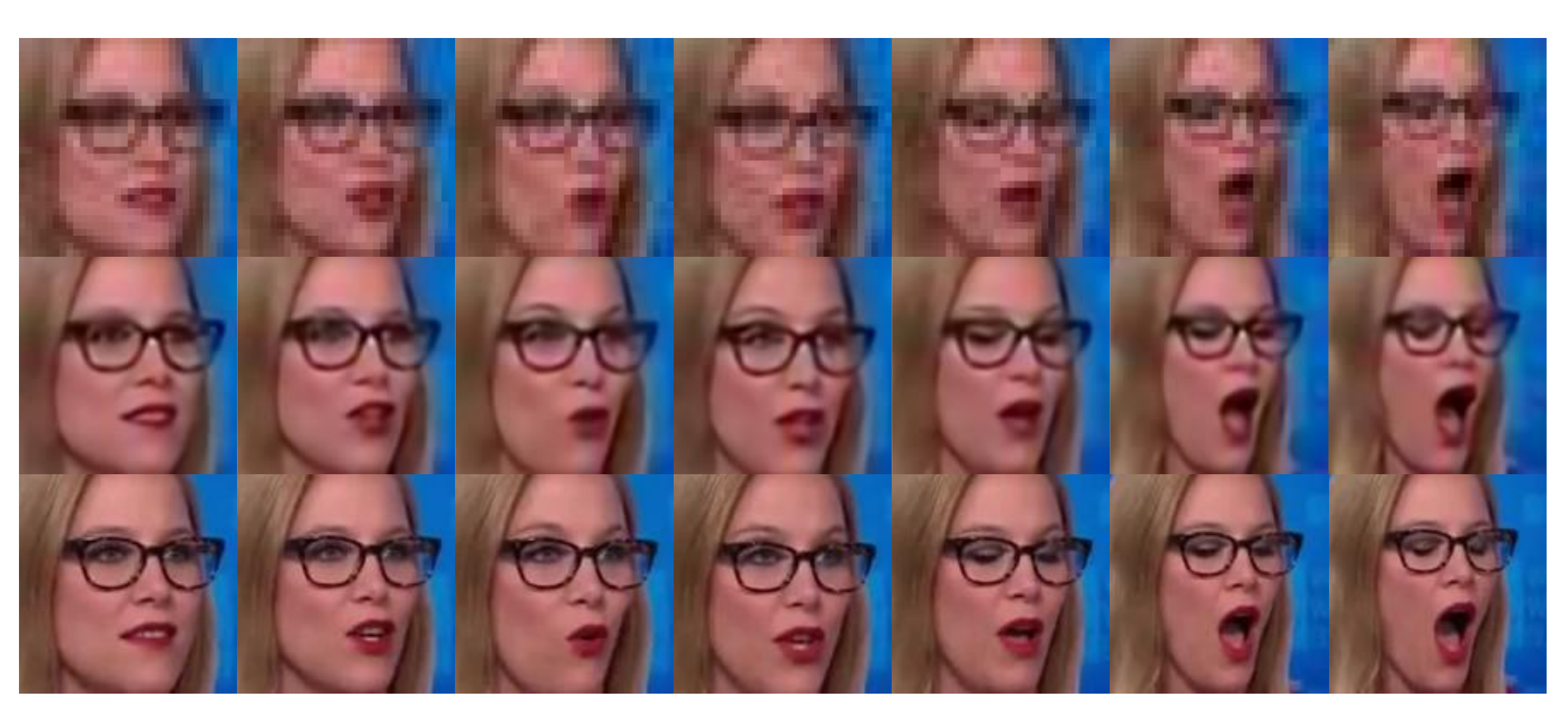}
	}
	\caption{Our method aims to generate high-resolution video frames (2nd row) from low-resolution ones (1st row, visualized using pixel duplication). The low-resolution frames (1st row) are downsampled from the corresponding groundtruth frames (3rd row) with noise and blur.}
	\label{fig:5}
\end{figure*}

\section{Introduction}

Image and video super-resolution (SR) is now an efficient method that could not only be widely used in many fields ranging from the medical and satellite imaging \cite{SESR,MSR} to the security and surveillance \cite{PSR,FD}, but also be used as an important preprocessing method for other machine vision tasks \cite{FR1,FR2,FR3,VFR}. Which has attracted much more attention in recent years. As a domain-specific of cross-modal learning \cite{yu2019deep,yu2018category}, SR technology aims to generate a High-Resolution (HR) image from a Low-Resolution (LR) input one, which guarantees the restoration of the low frequency information in the frequency band and predicts the high frequency information above the cutoff frequency. With the development of convolution neural networks, the deep learning based image super-resolution (SR) \cite{VDSR,ResNet} and face SR \cite{CFSR1,CFSR2,CFSR3} have received significant attention from the research community over the past few years, which achieve the state-of-the-art performances in terms of the peak signal-to-noise ratio (PSNR) and the structural similarity index (SSIM) \cite{SSIM}. However, these methods do not consider the temporal relationship between frames, that make its results in video super-resolution (VSR) is not satisfactory.

Recently, the multi-frame SR technology is proposed in the field of video super-resolution, in which the information of the motion relationship between input frames is utilized to improve the performance. It takes multiple LR frames as inputs and output HR frames by taking into account subpixel motions between the neighboring LR frames. Usually, most deep learning based VSR methods \cite{DESR,VSRNet,VESPCN,LIU,SPMC} follow a similar procedure that consists of two steps: the first step is the motion estimation and compensation, and the second one is an up-sampling process. The motion estimation and compensation are the key of these methods, it is a hard task, and especially when the complex motion or parallax appear across the neighboring frames. Caballero et.al.\cite{VESPCN} had proposed an efficient spatial transformer network to compensate the motion between frames fed to the SR network, but network performance decreases when the number of input frames exceeds 5. Jo et.al.\cite{VSR_DUF} had explored a method based on the dynamic upsampling filters estimation which avoid explicit motion compensation, but it is also hard to make full use of the video's motion information.

In addition, reducing the number of frames which input into the network at the same time, could effectively improve the network's ability to model the complex temporal dependencies\cite{FRVSR,RBPN}. However, the decrease in the number of input frames also means that the network be likely to receive less useful information, and the performance of the network will be further limited.

In the deep-learning based VSR methods abovementioned, the input images are connected in parallel and the network has not treated them discriminately, which limits the ability of networks to learn the useful information when the input frames were too large. This would lead to the performance of the network decrease too much as the input frames number increasing\cite{VESPCN,MMCNN}.

In this paper, we focus on the problem of the motion estimation and compensation of video super resolution, and propose a Motion-Adaptive Feedback Ceil (MAFC) and a novel network named Motion-Adaptive Feedback Network (MAFN) to improve the VSR performance. MAFC can sensitively capture the motion information between each frame and feed back to the network in an adaptive way, and MAFN can process each input image independently by the channel separation method which can meets the input requirements of MAFC. Furthermore, we apply the proposed model to super resolve videos. The experimental results demonstrate that our approach can be applied in the complex motion scenarios, and achieve state-of-the-art performance. Moreover, it can also effectively solve the problem of network performance degradation caused by excessive input frames.

The advantages of our model can be summarized as follows. 1). It makes full use of the motion relationship between different frames and avoids motion compensation operation. 2). It can enhance the connection between each frame features, and the different frame features can be treated discriminately. 3). The temporal dependency can be efficiently modelled by the network, and the performance of the network improves as the number of input frames increasing.

\section{Related Works}
Early works have made efforts on addressing the VSR problems by putting the motion between HR frames, the blurring process and the subsampling altogether into one framework and focused on solving for the sharp frames with an optimization \cite{VSR2}. Among these traditional methods, Protter et al. \cite{VSR_nonlocal} and Takeda et al. \cite{VSR_3D} avoided the motion estimation by employing nonlocal mean and 3D steering kernel regression. Liu and Sun \cite{VSR_Bayes} proposed a Bayesian approach to estimate HR video sequences, which can also compute the motion fields and blur kernels simultaneously.

Recently, with the rise of deep learning, various networks have been designed in video super-resolution field, such as early deep learning method BRCN \cite{BRCN} using recurrent neural networks to model long-term contextual information of temporal sequences. Specifically, they used bidirectional connection between video frames with three types of convolutions: the feedforward convolution for spatial dependency, the recurrent convolution for long-term temporal dependency, and the conditional convolution for long-term contextual information. Besides, Liao et al. \cite{DESR} proposed DESR, which reduces computational load for motion estimation by employing a noniterative framework. The SR drafts are generated by several hand-designed optical flow algorithms, leading to a deep network produce final results. Likewise, Kappeler et al. \cite{VSRNet} proposed VSRnet, which compensates motions in input LR frames by using a hand-designed optical flow algorithm as a preprocessing before being fed to a pretrained deep SR network. Caballero et al. \cite{VESPCN} proposed VESPCN,  which learns motions between input LR frames and improves HR frame reconstruction accuracy in real-time. Furthermore, this end-to-end deep network estimates the optical flow between input LR frames with a learned CNN to warp frames by a spatial transformer \cite{STN}, and produces an HR frame through another deep network.

\begin{figure*}[t]
	\captionsetup{aboveskip=10pt}
	\captionsetup{belowskip=10pt}
	\centering
	\scalebox{1.0}{
		\includegraphics[width=1.0\linewidth]{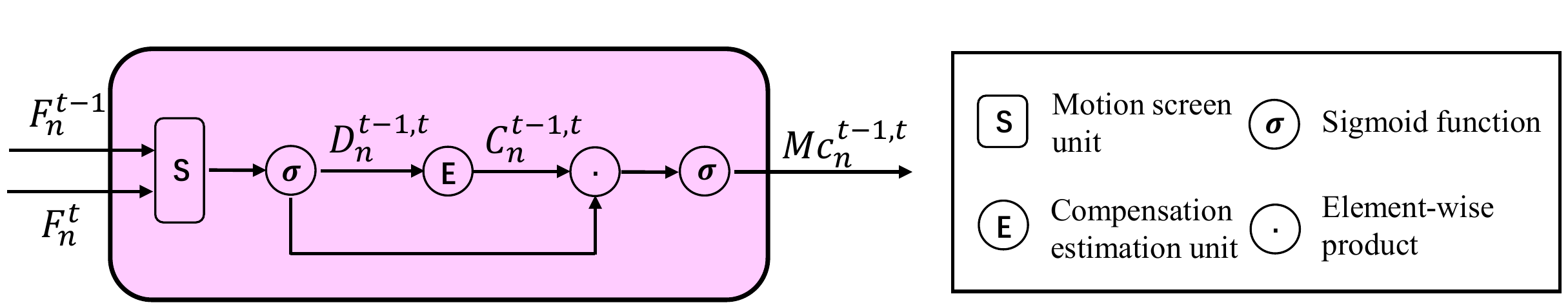}
	}
	\caption{The proposed motion-adaptive feedback cell (MAFC).}
	\label{fig:3}
\end{figure*}

Similar to the above methods, the work in \cite{LIU} also learns and compensates the motion between input LR frames. But after the motion compensation, they adaptively use the motion information in various temporal radius by temporal adaptive neural network. The network is composed of several SR inference branches for each different temporal radius, and the final output is generated by aggregating the outputs of all the branches. Tao et al. \cite{SPMC} used motion compensation transformer module from \cite{VESPCN} for the motion estimation, and proposed a subpixel motion compensation layer for simultaneous motion compensation and upsampling. For following SR network, an encoder-decoder style network with skip connections is used to accelerate the training and the ConvLSTM module is used since video is sequential data. Jo et.al \cite{VSR_DUF} designed a network which reconstructs image by generating the dynamic upsampling filters and a residual image. Sajjadi et.al \cite{FRVSR} proposed a frame-recurrent video super-resolution framework that uses the previously inferred HR estimate to super-resolve the subsequent frame. Li et.al \cite{FSTRN} optimized the structure of 3D convolution and proposes a fast VSR method. Inspired by the idea of back-projection, Haris et.al\cite{RBPN} integrated spatial and temporal contexts from continuous video frames using a recurrent encoder-decoder module.

\section{Method}

Our starting point is improving the ability of network to model the motion information from the video. We restrict our analysis to the motion compensation between each frame and do not further investigate potentially beneficial extensions such as recurrence \cite{DRCN} and residual learning \cite{VDSR}, width and depth of network \cite{RCAN} or different loss functions \cite{LapSRN,SRGAN}. After presenting an introduction of the MAFC in Sec. 3.1 and defining a novel network used for combining MAFC in Sec. 3.2, we justify our design choices in Sec. 3.3 and give details on the implementation and training procedure in Sec. 3.4 and 3.5, respectively.

\subsection{Motion-Adaptive Feedback Cell}

The motion compensation operations of the existing methods are performed directly on the input data, which can be seen as a form of preprocessing. Due to some complex motions are difficult to be modeled, the low-quality motion compensation methods will lead to the performance of network decrease drastically\cite{VESPCN,MMCNN}. What is more, the temporal dependencies among input frames may become too complex for networks to learn useful information, and act as noise degrading their performance\cite{VESPCN}.

Considering the problem mentioned above, we advise a motion feedback mechanism between each frame feature and propose a Motion-Adaptive Feedback Cell (MAFC), which can update the current frame features adaptively according to the difference between its neighboring frames. The overview of the proposed MAFC is shown in Fig.\ref{fig:3}.

As shown in Fig.\ref{fig:3}, two input $F^{t}_{n}$ and $F^{t-1}_{n}$ are the feature maps from different frames but with the same convolutional receptive field, and the output $Mc^{t,t-1}_{n}$ is the motion compensation information of the input frames. The first step in MAFC is to throw away the redundant information from the cell state by a motion screen unit (MSU), and normalize it between 0 and 1. Next, we use the updated information $D^{t,t-1}_{n}$ to infer the candidate values of motion compensation by a compensation estimation unit (CEU). Finally, we combine the two strategies  to create the final cell state and use them as motion compensation features $Mc^{t,t-1}_{n}$. Among them :

\begin{equation}
\begin{aligned}
D_n^{t,t - 1} = \sigma ({W_{m1}}[F_n^t, F_n^{t{\rm{-1}}}] + b_{m1}),
\end{aligned}
\label{Eq:1}
\end{equation}

\begin{equation}
\begin{aligned}
C_n^{t,t - 1} = {W_{m2}}[D_n^{t,t-1}] + b_{m2},    
\end{aligned}
\label{Eq:2}
\end{equation}

\begin{equation}
\begin{aligned}
Mc^{t,t-1}_{n} = \sigma(C_n^{t,t - 1} * D_n^{t,t - 1}),    
\end{aligned}
\label{Eq:3}
\end{equation}

The motion screen unit is made by a reduction operating followed by a convolution layer and a sigmoid layer, and the compensation estimation unit similarly consists of two convolution layers and a sigmoid layer. Where $\sigma$ is a sigmoid layer, $W_{m1}$ and $W_{m2}$ are the convolution parameters of MSU and CEU.

\begin{figure*}[t]
	\captionsetup{belowskip=-0pt}
	\centering
	\scalebox{0.95}{
	\includegraphics[width=1.0\linewidth]{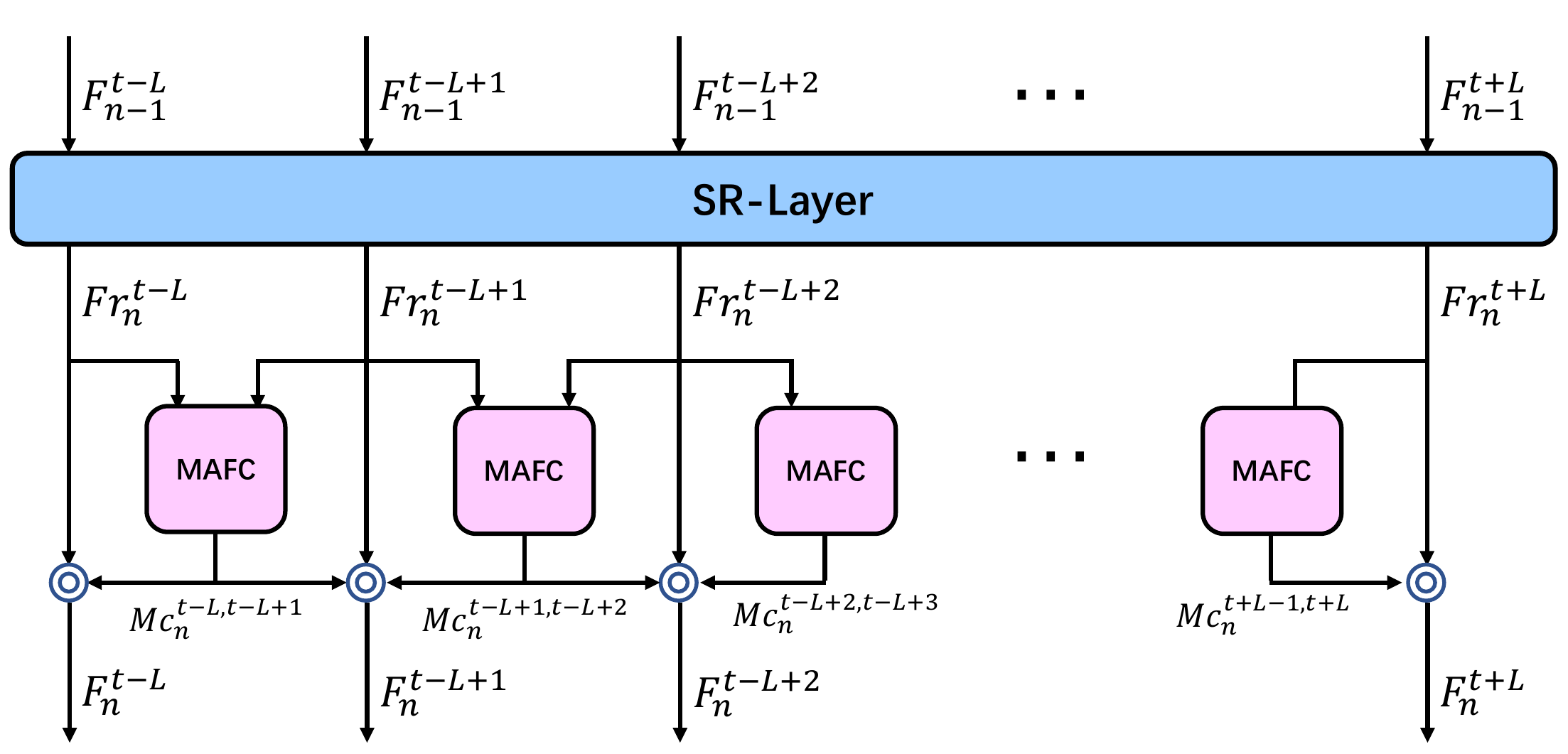}
	}
	\caption{Network architectures for one mid-block of MAFN, named MAFB.}
	\label{fig:2}
\end{figure*}

In short, MSU is the initial screening of motion information, while CEU is the adaptive enhancement and restrain of the screened motion. Generally, the motion existing in video is very complex, such as the motion of image background, rigid motion of the whole face, non-rigid motion of facial expression and so on. The contribution of these motions to image reconstruction is different. The goal of the MSU screening process is to filter out unimportant movements such as background movements. The CEU is work to estimate the importance degree of the remaining multi-type motions $ D_n^{t,t - 1} $ after MSU's screening, and generate their corresponding weight coefficients $ C_n^{t,t - 1} $ adaptively. Then, at the end of the MAFC, The multiplication of $ D_n^{t,t - 1} $ and $ C_n^{t,t - 1} $ accomplishes the function for multi-type motions's enhance or restrain, and feedback to the network more clear and concise motion compensation features.

\subsection{Network Design}

Given a low-resolution, noisy and blurry video ${X_t}$, the goal of the VSR is to estimate a high-resolution, noise-free and blur-free version ${\widehat Y_t}$. The LR frames ${X_t}$ are downsampled from the corresponding groundtruth (GT) frames ${Y_t}$ with noise and blur, where $t$ denotes the time step. With the VSR network G and the network parameters $\theta$, the VSR problem is defined as:

\begin{equation}
\begin{aligned}
{\widehat Y_t} = {G_{\theta}}(X_{t-L : t+L}),
\end{aligned}
\label{Eq:4}
\end{equation}

where $L$ is the temporal radius and the input frames number T for the network is $2L+1$.

\begin{figure*}
	\captionsetup{aboveskip=10pt}
	\captionsetup{belowskip=10pt}
	\includegraphics[width=1.0\linewidth]{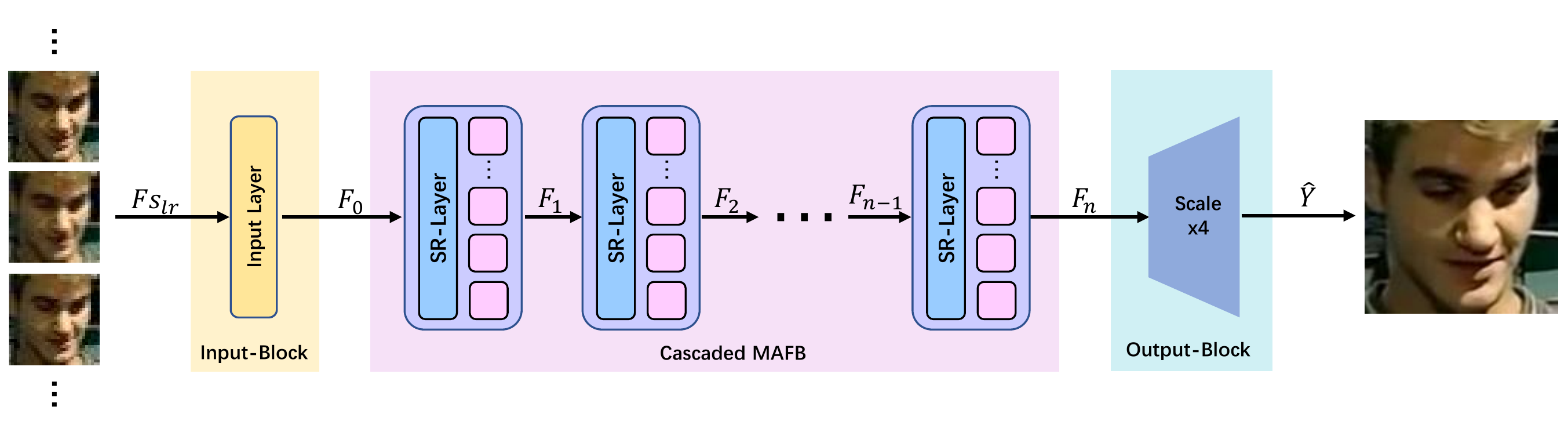}
	\caption{Pipeline of our proposed MAFN model.}
	\label{fig:1}
\end{figure*}

MAFC requires the two inputs  from different frames to work properly. Unfortunately, none of existing deep learning based VSR methods work in this way. Thus, our MAFC cannot be carried out in the existing network structures. To address this problem, we simply conduct a lightweight network as shown in Fig.\ref{fig:1}, which consists of three parts: an Input-Block to map an input frames ${Fs_{lr}}$ into the deep features, a Mid-Block to convert the features to the more complete facial presentation features, and a Output-Block to produce the output image ${\widehat Y}$ from the facial presentation features. The detail of mid-block is shown in Fig.\ref{fig:2}. Besides, it is worth noting that each mid-layer has the same structure.

For the last layer output $F_{n-1}$, we first update the representation features of each input by a simple SR block and obtain the $Fr_{n}$. Then, the adjacent features are used as the inputs and sent to the MAFC in pairs, leading to the corresponding motion compensation information $Mc_{n}$. It should be noted that each MAFC has the same model parameters in this layer, and that is MAFC utilizes a weight sharing way to ensure the fairness of motion compensation operation between any frames. Finally, we combine the motion compensation information with the newly obtained representation features to obtain the final output $F_{n}$. Among them:

\begin{equation}
\begin{aligned}
F_{n-1} = [F^{t-L}_{n-1},..., F^{t}_{n-1},...,F^{t+L}_{n-1}], 
\end{aligned}
\label{Eq:5}
\end{equation}

\begin{equation}
\centering
\begin{aligned}
&Fr^{t}_{n} = \sigma ({W_{f}}[F^{t}_{n-1}] + b_{f}), \\[2mm]
Mc&^{t,t+1}_{n} = MAFC(Fr^{t}_{n},  Fr^{t}_{n+1}),
\end{aligned}
\label{Eq:6}
\end{equation}

\begin{equation}
\begin{aligned}
&F^{t}_{n} = [Mc^{t-1,t}_{n}, Fr^{t}_{n}, Mc^{t,t+1}_{n}], \\[2mm]
&F_{n} = [F^{t-L}_{n},..., F^{t}_{n},...,F^{t+L}_{n}], 
\end{aligned}
\label{Eq:7}
\end{equation}

where $W_{f}$ is the parameters of SR block. For each updated representation feature $Fr^{t}_{n}$, we combine its adjacent motion compensation information and itself as the final output.  Then, let $k3$ denote that the convolution kernel size is 3, $s1$ denotes that the stride is 1 and $d16$ denotes that the number of feature channels is 16. The Input-Layer is made by a $k3s1d16$ followed by a relu function, SR-Block is two $k3s1d16$ and a relu function, and the Output-Block architecture is: $k3s1d128$, $PixelShuffle$\_x2, $k3s1d64$, $PixelShuffle$\_x2, $k3s1d1$.

\subsection{Why Does MAFC Work Better?}

A short answer is that it could more efficient utilize the motion information against common motion compensation operation. The most methods rely heavily on the accuracy of motion estimation and compensation. However the complex motions are difficult to model, it can introduce adverse effects if not handled properly. Specifically, while motion compensation operation such as the STN \cite{STN} is essential pieces in almost all the state-of-the-art VSR models \cite{DESR,VSRNet,VESPCN,LIU,SPMC,FRVSR}, they tend to increase the spatial homogeneity information through affine transformation and interpolation. Therefore, it may ignore the problem of model complexity caused by breaking spatial consistency. 

Another aspect of the limitations is that the fusion process of the continuous frame is carried out in a way of weighted sum. As a result, it hard to make full use of the variation and correlation between each frame to model video motion. When too many frames are sent into the network, the performance of the network will decline significantly.

The advantages of our method can be reflected in two aspects. On the one hand, our network map the separated images of each frame into the same feature space by the convolution layer with weight sharing. In this space, the difference between each group of features represents each frame’s motion. MAFC could more efficient and intuitive to extract the variations between each frame features and feed back them to the network. the complexity of the network's temporal dependence modeling has been greatly reduced. On the other hand, because of the MAFC works after each SR-Layer, motion imformation with different convolution receptive fields could be simultaneously utilized by the network. So that the network has a richer source of information for modeling temporal dependencies. In the subsequent experiments, we further prove that our method has better modeling ability for the complex time dependence of video.

\subsection{Implementation Details}
\subsubsection{Dataset}

We conduct experiments on VoxCeleb dataset. It contains over 1 million utterances for 6,112 celebrities, extracted from videos uploaded to YouTube, which provides the sequences of tracked faces in the form of bounding boxes. 

\begin{table}[h]
	\captionsetup{belowskip=-5pt}
	\captionsetup{aboveskip=10pt}
	\small
	\centering
	\begin{tabular}{p{2cm}p{1.5cm}p{1.5cm}p{1.5cm}}
		\toprule
		\multirow{2}{*}{Dataset} & \multicolumn{3}{c}{VoxCeleb} \\
		& objects & sequences & frames  \\
		\midrule
		Training & 100  & 3884 & 776640 \\
		Validation & 5 & 10 & 2144\\
		Testing & 18 & 697 & 139368\\
		\bottomrule
	\end{tabular}
	\caption{Datasets used in facial video super-resolution.}
	\label{tab:1}
\end{table}

Here we select 3884 video sequences of 100 people for training, 10 video sequences of 5 people for verification and 697 sequences of 18 people for testing. For each sequence, we compute a box enclosing the faces from all frames and use it to crop face images from the original video. All face images are resized to $128\times128$. Table.\ref{tab:1} presents the split of training, validation and testing sets.

\subsubsection{Degradation models}

Considering the influence of a variety of adverse factors in the image acquisition processing, the obtained image may have some problems such as noise, blur and low resolution at the same time. The image degradation model can be approximated as :

\begin{equation}
\begin{aligned}
{X_t} = {D_t}{B_t}{Y_t} + {Z_t},
\end{aligned}
\label{Eq:8}
\end{equation}

where $D_t$ and $B_t$ is the downsampling and fuzzy matrix, $Z_t$ is the additional noise. Our LR inputs are  generated from HR frames according to the above image degradation model. We first blur HR image by Gaussian kernel of size $7\times7$ with standard deviation $1.6$,  bicubic downsample HR image with scaling factor $4$, and then add Gaussian noise with noise level $5$ \cite{RDN}.

\begin{table*}
	\captionsetup{belowskip=0pt}
	\captionsetup{aboveskip=10pt}
	\centering
	\fontsize{9}{12}\selectfont
	\scalebox{1.3}{
		\begin{tabular}{p{2cm}p{1.2cm}p{1.2cm}p{1.2cm}p{1.2cm}p{1.2cm}p{1.2cm}} 
			\hline
			\multirow{2}{*}{Methods} & \multicolumn{2}{c}{$T=3$} & \multicolumn{2}{c}{$T=5$} &\multicolumn{2}{c}{$T=7$} \\
			
			& PSRN & SSIM & PSRN & SSIM & PSRN & SSIM \\
			\hline
			Bicubic & 29.95 & 0.8416 & 29.95 & 0.8416 & 29.95 & 0.8416 \\
			
			DESR & 32.19 & 0.8929 & 32.30 & 0.8953 & 32.09 & 0.8929\\
			
			VESPCN & 33.07 & 0.9097 & 33.14 & 0.9112 & 32.79 & 0.9055\\
			
			LIU et.al & 32.70 & 0.9033 & 32.82 & 0.9063 & 32.66 & 0.9033\\
			
			SPMC & 33.03 & 0.9066 & 33.23 & 0.9099 & 33.44 & 0.9132\\
			
			FRVSR & 33.26 & 0.9105 & 33.42 & 0.9129 & 33.53 & 0.9147\\
			
			VSR\_DUF & {\bf 34.38} & {\bf 0.9290} &  34.14 & 0.9245 & 33.82 & 0.9214\\
			
			FSTRN & 32.96 & 0.9059 & 33.11 & 0.9089 & 33.07 & 0.9085\\
			
			RBPN & 33.16 & 0.9084 & 33.67 & 0.9158 & 33.91 & 0.9232\\
			\hline
			MAFN  & 34.15 & 0.9237 & {\bf 34.59} & {\bf 0.9279} & {\bf 34.81} & {\bf 0.9318}\\
			\hline
		\end{tabular}
	}
	\caption{Performance of facial video hallucination on the testing sets}
	\label{tab:2}
\end{table*}

\subsection{Training Procedure}

The pipeline of our network structure is shown in Fig.3, named Motion-Adaptive Feedback Network MAFN, which is a flexible network. For our experiments, the network has one input layer, one output layer and seven SR blocks, which only consists of two convolution layers and one sigmoid layer. Furthermore, the loss function of MAFN is:

\begin{equation}
\begin{aligned}
L_G(\theta ) = \frac{1}{M}\sum\limits_{i = 1}^M \{ \left\| {Y_t} - {\widehat Y_t} \right\| \}, 
\end{aligned}
\label{Eq:9}
\end{equation}
where $M$ is the number of training images. We implement our model by using the pytorch environment, and optimize our network by Adam with back propagation. The momentum parameter is set to 0.1, weight decay is set to $2 \times {10^{{\rm{ - }}4}}$, and the initial learning rate is set to $1 \times {10^{{\rm{ - }}3}}$ and be divided a half every 10 epochs. Batchsize is set to 16. Training a MAFN on VoxCeleb dataset generally takes 10 hours with one Titan X Pascal GPU. For assessing the quality of SR results, we employ two objective image quality assessment metrics: Peak Signal to Noise Ratio (PSNR) and structural similarity (SSIM). All metrics are performed on the Y-channel (YCbCr color space) of super-resolved images.

\section{Evaluation}

In order to demonstrate the effect of our proposed MAFC, we first compare the proposed network with other state-of-the-art methods. Furthermore,  we also investigate the impact of model architecture and input frames number on the performance.

\subsection{Comparisons with State-of-the-Arts}

{\bf Quantitative comparisons}

We compare our proposed MAFN with the state-of-the-art VSR methods, including DESR \cite{DESR}, VESPCN \cite{VESPCN}, LIU et.al \cite{LIU}, SPMC \cite{SPMC}, FRVSR \cite{FRVSR}, VSR\_DUF \cite{VSR_DUF} FSTRN \cite{FSTRN} and RBPN \cite{RBPN}. For fair comparison, we train all models with the same training set. To demonstrate the ability of each method to model complex temporal dependencies, we use different numbers of input frames to train and test the network, where $T \in \{ 3,5,7\}$.

Quantitative comparison with other state-of-the-art VSR methods is shown in Table.\ref{tab:2}. In general, VSR\_DUF \cite{VSR_DUF} nearly achieves the best performance except our method, but the performance decreases significantly as the number of input frames increases. We think the reason is this method has no explicitly motion estimation and compensation operation, which making it difficult for the network to model the complex dependencies between frames. In addition, due to a lightweight structure, our method does not get the highest performance at $T=3$. Then, our method achieves the best performance when $T=5$, and the performance of all methods except VSR\_DUF has been improved. It can be seen that motion compensation operation could enhance the network's modeling ability for complex motion. 

Specifically, when $T=7$, the performance of SPMC, FRVSR, RBPN and ours methods increased, but other methods decreased. The common feature of FRVSR and RBPN is that the network input only contains two frames of images at the same time. As for the SPMC, we modified the way the network reads in the image and made it consistent with the FRVSR. It is found that the network input with fewer images could effectively increase the network's ability to model temporal dependence. The advantage of this method is that the performance of the network can increase with the increase of the input frames number, but the lack of input also could lead to the performance of the network is hard to further improve. Benefit from the ability of MAFC to utilize inter-frame motion information, our method achieves excellent performance at T=7. Moreover, compared with T=5, the performance of the MAFN has a significant increase.

{\bf Qualitative results}

\begin{figure*}
	\captionsetup{belowskip=0pt}
	\captionsetup{aboveskip=5pt}
	\centering
	\scalebox{0.8}{
	\includegraphics[width=1.0\linewidth]{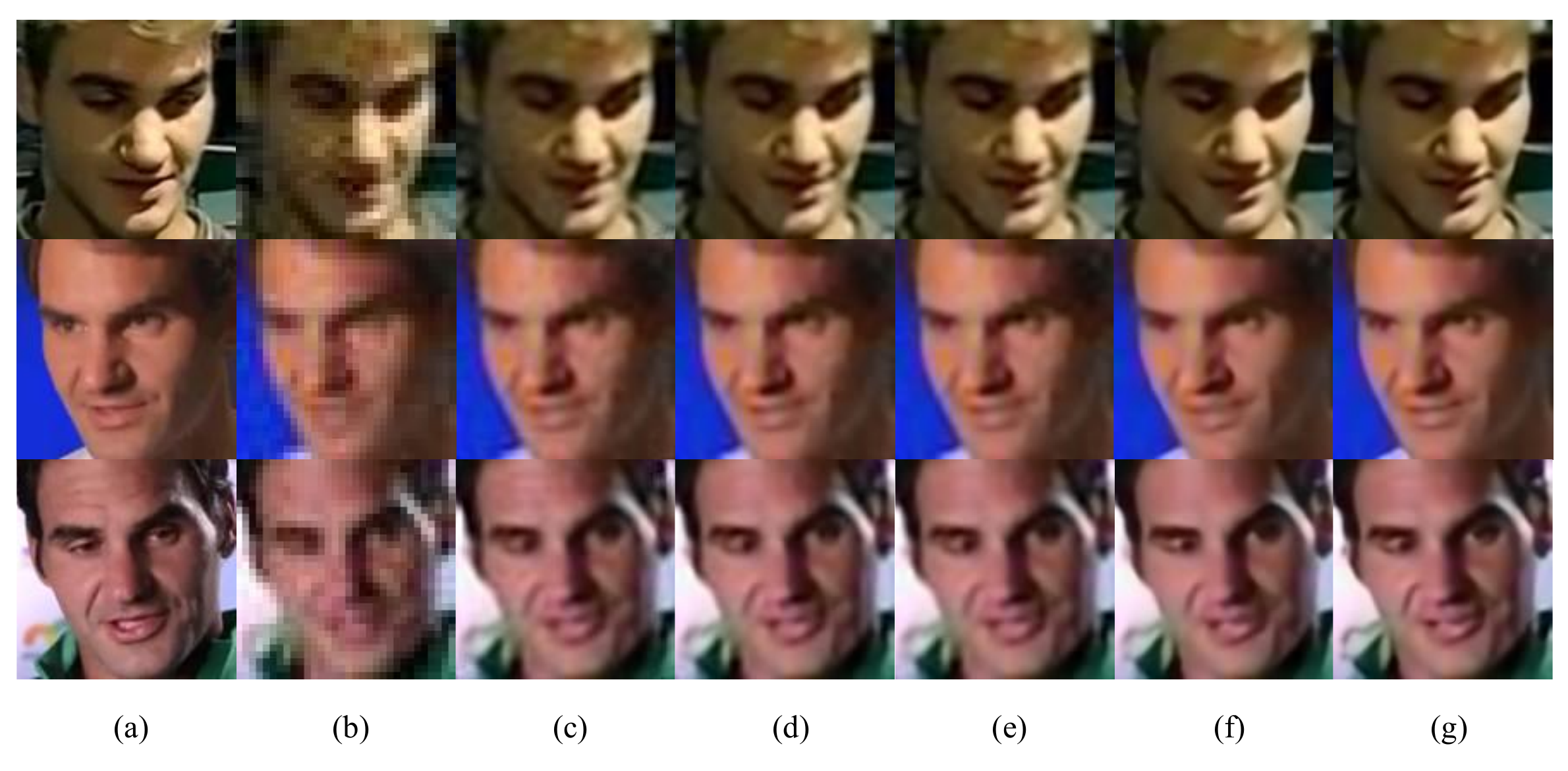}
	}
	\caption{Visual evaluation on scale 4. (a) Original HR images. (b) Input LR images. (c) Results of Caballero $et$ $al.'s$ method (VESPCN). (d) Results of Sajjadi $et$ $al.'s$ method (FRVSR). (e) Results of Jo $et$ $al.'s$ method (VSR\_DUF). (f) Results of Haris $et$ $al.'s$ method (RBPN). (g) Results of our MAFN.}
	\label{fig:4}
\end{figure*}

A qualitative comparison between our method and other SR methods are shown in Figure. \ref{fig:4}. There are three face images, each of which is reconstructed from seven consecutive frames. The super-resolved results from our method tend to be more appealing and clearer than those from other methods especially on the mouth and eyes.

In Figure. \ref{fig:5}, we show more example results from the voxceleb dataset. This method is applied to many scenes and has high image fidelity.

\subsection{Ablation Study}

We conduct the ablation study on our proposed network to demonstrate the effects of our methods. Since our network is a really simple network which is similar to the Fast Super Resolution Convolution Neural Network \cite{FSRCNN}, it doesn't merit any additional discussion here. In this section, we mainly make a detailed discussion on how to improve the performance with MAFC. We conduct 4 experiments to estimate the basic network, signal screen uint, compensation estimation uint, and MAFC, respectively. Specifically, by removing the MAFC from our MAFN, the remaining parts constitute the first network, named ‘BasicNet v1’. The second network, named ‘BasicNet v2’, has the same structure as ‘MAFN’ except that the MAFC retains only the signal screen uint. in the same way, the third network ‘BasicNet v3’ is the MAFC retains only the compensation estimation uint. In this part, we study the effects of different networks.  For fairly comparison, the differences between the four networks are only the part of MAFC and we train all those models with other same implementation details.

\begin{table}[h]
	\captionsetup{aboveskip=10pt}
	\small
	\centering
	\begin{tabular}{p{0.6cm}p{1.2cm}p{1.2cm}p{1.2cm}p{1.2cm}}
		\toprule
		{T} & {BasicNet v1} & {BasicNet v2} & {BasicNet v3} & {MAFN}  \\
		\midrule
		3 & 32.48 & 33.89 & 33.82  & {\bf 34.15} \\
		5 & 32.64  & 34.20 & 33.11 & {\bf 34.59} \\
		7 & 32.55 & 34.37 & 33.29  & {\bf 34.81}\\
		\bottomrule
	\end{tabular}
	\caption{Ablation study on effects of MAFC.}
	\label{tab:3}
\end{table}

Table.\ref{tab:3} shows the results of different network structures. It can be seen that: (1) Compared to other networks, the basic network (BasicNet v1) has lower performance and the increase of the number of input frames does not significantly improve its results. (2) BasicNet v2 and BasicNet v3 are both can achieve good performance. It can be seen that the subtraction followed by convolution operation and the parallel followed by convolution operation can achieve motion compensation for input two-frame features. (3) The model using two uint (signal screen and compensation estimation) achieves the best performance, which indicates that more rich and distinct motion compensation information brings more improvement.

\section{Conclusion}

In this paper, we propose a Motion-Adaptive Feedback Ceil (MAFC) and a novel network named Motion-Adaptive Feedback Network (MAFN) for video super-resolution. The key contribute of this paper is that, we find the shortcoming of the current VSR method based on the motion estimation and compensation, and put forward an adaptive feedback method to deal with its drawback, and obtain the satisfactory results. The advantage of MAFC is it can efficiently extract the differences between each frame features intuitively, and capture the differences of each frame in each level representation space, which makes the motion information between each frame image could be learned more sufficiently. Extensive experiments show that MAFC significantly outperforms state-of-the-arts. Thus, we believe this motion adaptive feedback strategy could be more widely applicable in practice, and it is readily used to other machine vision problems such as video deblurring, compression artifact removal and even optical flow learning.

\section{Acknowledgement}
This work was supported in part by the National Natural Science Foundation of China under Grant Grant 61922066, Grant 61876142, Grant 61671339, Grant 61772402, Grant U1605252, Grant 61432014, in part by the National Key Research and Development Program of China under Grant 2016QY01W0200 and Grant 2018AAA0103202, in part by the National High-Level Talents Special Support Program of China under Grant CS31117200001, in part by the Fundamental Research Funds for the Central Universities under Grant JB190117, in part by the Xidian University-Intellifusion Joint Innovation Laboratory of Artificial Intelligence, in part by the Innovation Fund of Xidian University.

\bibliographystyle{IEEEtran}
\bibliography{MAFC}

\begin{thebibliography}{10}
\providecommand{\url}[1]{#1}
\csname url@samestyle\endcsname
\providecommand{\newblock}{\relax}
\providecommand{\bibinfo}[2]{#2}
\providecommand{\BIBentrySTDinterwordspacing}{\spaceskip=0pt\relax}
\providecommand{\BIBentryALTinterwordstretchfactor}{4}
\providecommand{\BIBentryALTinterwordspacing}{\spaceskip=\fontdimen2\font plus
\BIBentryALTinterwordstretchfactor\fontdimen3\font minus
  \fontdimen4\font\relax}
\providecommand{\BIBforeignlanguage}[2]{{%
\expandafter\ifx\csname l@#1\endcsname\relax
\typeout{** WARNING: IEEEtran.bst: No hyphenation pattern has been}%
\typeout{** loaded for the language `#1'. Using the pattern for}%
\typeout{** the default language instead.}%
\else
\language=\csname l@#1\endcsname
\fi
#2}}
\providecommand{\BIBdecl}{\relax}
\BIBdecl

\bibitem{SESR}
M.~W. Thornton, P.~M. Atkinson, and D.~Holland, ``Sub-pixel mapping of rural
  land cover objects from fine spatial resolution satellite sensor imagery
  using super-resolution pixel-swapping,'' \emph{International Journal of
  Remote Sensing}, vol.~27, no.~3, pp. 473--491, 2006.

\bibitem{MSR}
W.~Shi, J.~Caballero, C.~Ledig, X.~Zhuang, W.~Bai, K.~Bhatia, A.~M. S.~M.
  de~Marvao, T.~Dawes, D.~O~Regan, and D.~Rueckert, ``Cardiac image
  super-resolution with global correspondence using multi-atlas patchmatch,''
  in \emph{International Conference on Medical Image Computing and
  Computer-Assisted Intervention}.\hskip 1em plus 0.5em minus 0.4em\relax
  Springer, 2013, pp. 9--16.

\bibitem{PSR}
W.~W. Zou and P.~C. Yuen, ``Very low resolution face recognition problem,''
  \emph{IEEE Transactions on image processing}, vol.~21, no.~1, pp. 327--340,
  2011.

\bibitem{FD}
K.~Zhang, Z.~Zhang, H.~Wang, Z.~Li, Y.~Qiao, and W.~Liu, ``Detecting faces
  using inside cascaded contextual cnn,'' in \emph{Proceedings of the IEEE
  International Conference on Computer Vision}, 2017, pp. 3171--3179.

\bibitem{FR1}
Z.~Li, W.~Liu, D.~Lin, and X.~Tang, ``Nonparametric subspace analysis for face
  recognition,'' in \emph{2005 IEEE Computer Society Conference on Computer
  Vision and Pattern Recognition (CVPR'05)}, vol.~2.\hskip 1em plus 0.5em minus
  0.4em\relax IEEE, 2005, pp. 961--966.

\bibitem{FR2}
Y.~Xiong, W.~Liu, D.~Zhao, and X.~Tang, ``Face recognition via archetype hull
  ranking,'' in \emph{Proceedings of the IEEE International Conference on
  Computer Vision}, 2013, pp. 585--592.

\bibitem{FR3}
H.~Wang, Y.~Wang, Z.~Zhou, X.~Ji, D.~Gong, J.~Zhou, Z.~Li, and W.~Liu,
  ``Cosface: Large margin cosine loss for deep face recognition,'' in
  \emph{Proceedings of the IEEE Conference on Computer Vision and Pattern
  Recognition}, 2018, pp. 5265--5274.

\bibitem{VFR}
W.~Liu, Z.~Li, and X.~Tang, ``Spatio-temporal embedding for statistical face
  recognition from video,'' in \emph{European Conference on Computer
  Vision}.\hskip 1em plus 0.5em minus 0.4em\relax Springer, 2006, pp. 374--388.

\bibitem{yu2019deep}
Y.~Yu, S.~Tang, F.~Raposo, and L.~Chen, ``Deep cross-modal correlation learning
  for audio and lyrics in music retrieval,'' \emph{ACM Transactions on
  Multimedia Computing, Communications, and Applications (TOMM)}, vol.~15,
  no.~1, p.~20, 2019.

\bibitem{yu2018category}
Y.~Yu, S.~Tang, K.~Aizawa, and A.~Aizawa, ``Category-based deep cca for
  fine-grained venue discovery from multimodal data,'' \emph{IEEE transactions
  on neural networks and learning systems}, vol.~30, no.~4, pp. 1250--1258,
  2018.

\bibitem{VDSR}
J.~Kim, J.~Kwon~Lee, and K.~Mu~Lee, ``Accurate image super-resolution using
  very deep convolutional networks,'' in \emph{Proceedings of the IEEE
  conference on computer vision and pattern recognition}, 2016, pp. 1646--1654.

\bibitem{ResNet}
K.~He, X.~Zhang, S.~Ren, and J.~Sun, ``Deep residual learning for image
  recognition,'' in \emph{Proceedings of the IEEE conference on computer vision
  and pattern recognition}, 2016, pp. 770--778.

\bibitem{CFSR1}
L.~Wei, D.~Lin, and X.~Tang, ``Hallucinating faces: Tensorpatch
  super-resolution and coupled residue compensation,'' in \emph{Computer Vision
  and Pattern Recognition, 2005. CVPR 2005. IEEE Computer Society Conference
  on}, 2005.

\bibitem{CFSR2}
L.~Wei, X.~Tang, and J.~Liu, ``Bayesian tensor inference for sketch-based
  facial photo hallucination,'' in \emph{IJCAI 2007, Proceedings of the 20th
  International Joint Conference on Artificial Intelligence, Hyderabad, India,
  January 6-12, 2007}, 2007.

\bibitem{CFSR3}
K.~Zhang, Z.~Zhang, C.-W. Cheng, W.~H. Hsu, Y.~Qiao, W.~Liu, and T.~Zhang,
  ``Super-identity convolutional neural network for face hallucination,'' in
  \emph{Proceedings of the European Conference on Computer Vision (ECCV)},
  2018.

\bibitem{SSIM}
Z.~Wang, A.~C. Bovik, H.~R. Sheikh, E.~P. Simoncelli \emph{et~al.}, ``Image
  quality assessment: from error visibility to structural similarity,''
  \emph{IEEE transactions on image processing}, vol.~13, no.~4, pp. 600--612,
  2004.

\bibitem{DESR}
R.~Liao, X.~Tao, R.~Li, Z.~Ma, and J.~Jia, ``Video super-resolution via deep
  draft-ensemble learning,'' in \emph{Proceedings of the IEEE International
  Conference on Computer Vision}, 2015, pp. 531--539.

\bibitem{VSRNet}
A.~Kappeler, S.~Yoo, Q.~Dai, and A.~K. Katsaggelos, ``Video super-resolution
  with convolutional neural networks,'' \emph{IEEE Transactions on
  Computational Imaging}, vol.~2, no.~2, pp. 109--122, 2016.

\bibitem{VESPCN}
J.~Caballero, C.~Ledig, A.~Aitken, A.~Acosta, J.~Totz, Z.~Wang, and W.~Shi,
  ``Real-time video super-resolution with spatio-temporal networks and motion
  compensation,'' in \emph{Proceedings of the IEEE Conference on Computer
  Vision and Pattern Recognition}, 2017, pp. 4778--4787.

\bibitem{LIU}
D.~Liu, Z.~Wang, Y.~Fan, X.~Liu, Z.~Wang, S.~Chang, and T.~Huang, ``Robust
  video super-resolution with learned temporal dynamics,'' in \emph{Proceedings
  of the IEEE International Conference on Computer Vision}, 2017, pp.
  2507--2515.

\bibitem{SPMC}
X.~Tao, H.~Gao, R.~Liao, J.~Wang, and J.~Jia, ``Detail-revealing deep video
  super-resolution,'' in \emph{Proceedings of the IEEE International Conference
  on Computer Vision}, 2017, pp. 4472--4480.

\bibitem{VSR_DUF}
Y.~Jo, S.~Wug~Oh, J.~Kang, and S.~Joo~Kim, ``Deep video super-resolution
  network using dynamic upsampling filters without explicit motion
  compensation,'' in \emph{Proceedings of the IEEE Conference on Computer
  Vision and Pattern Recognition}, 2018, pp. 3224--3232.

\bibitem{FRVSR}
M.~S. Sajjadi, R.~Vemulapalli, and M.~Brown, ``Frame-recurrent video
  super-resolution,'' in \emph{Proceedings of the IEEE Conference on Computer
  Vision and Pattern Recognition}, 2018, pp. 6626--6634.

\bibitem{RBPN}
M.~Haris, G.~Shakhnarovich, and N.~Ukita, ``Recurrent back-projection network
  for video super-resolution,'' in \emph{Proceedings of the IEEE Conference on
  Computer Vision and Pattern Recognition}, 2019, pp. 3897--3906.

\bibitem{MMCNN}
Z.~Wang, P.~Yi, K.~Jiang, J.~Jiang, Z.~Han, T.~Lu, and J.~Ma, ``Multi-memory
  convolutional neural network for video super-resolution,'' \emph{IEEE
  Transactions on Image Processing}, vol.~28, no.~5, pp. 2530--2544, 2018.

\bibitem{VSR2}
Z.~Ma, R.~Liao, X.~Tao, L.~Xu, J.~Jia, and E.~Wu, ``Handling motion blur in
  multi-frame super-resolution,'' in \emph{Proceedings of the IEEE Conference
  on Computer Vision and Pattern Recognition}, 2015, pp. 5224--5232.

\bibitem{VSR_nonlocal}
M.~Protter, M.~Elad, H.~Takeda, and P.~Milanfar, ``Generalizing the
  nonlocal-means to super-resolution reconstruction,'' \emph{IEEE Transactions
  on image processing}, vol.~18, no.~1, pp. 36--51, 2008.

\bibitem{VSR_3D}
H.~Takeda, P.~Milanfar, M.~Protter, and M.~Elad, ``Super-resolution without
  explicit subpixel motion estimation,'' \emph{IEEE Transactions on Image
  Processing}, vol.~18, no.~9, pp. 1958--1975, 2009.

\bibitem{VSR_Bayes}
C.~Liu and D.~Sun, ``On bayesian adaptive video super resolution,'' \emph{IEEE
  transactions on pattern analysis and machine intelligence}, vol.~36, no.~2,
  pp. 346--360, 2013.

\bibitem{BRCN}
Y.~Huang, W.~Wang, and L.~Wang, ``Bidirectional recurrent convolutional
  networks for multi-frame super-resolution,'' in \emph{Advances in Neural
  Information Processing Systems}, 2015, pp. 235--243.

\bibitem{STN}
M.~Jaderberg, K.~Simonyan, A.~Zisserman \emph{et~al.}, ``Spatial transformer
  networks,'' in \emph{Advances in neural information processing systems},
  2015, pp. 2017--2025.

\bibitem{FSTRN}
S.~Li, F.~He, B.~Du, L.~Zhang, Y.~Xu, and D.~Tao, ``Fast spatio-temporal
  residual network for video super-resolution,'' \emph{arXiv preprint
  arXiv:1904.02870}, 2019.

\bibitem{DRCN}
J.~Kim, J.~Kwon~Lee, and K.~Mu~Lee, ``Deeply-recursive convolutional network
  for image super-resolution,'' in \emph{Proceedings of the IEEE conference on
  computer vision and pattern recognition}, 2016, pp. 1637--1645.

\bibitem{RCAN}
Y.~Zhang, K.~Li, K.~Li, L.~Wang, B.~Zhong, and Y.~Fu, ``Image super-resolution
  using very deep residual channel attention networks,'' in \emph{Proceedings
  of the European Conference on Computer Vision (ECCV)}, 2018, pp. 286--301.

\bibitem{LapSRN}
W.-S. Lai, J.-B. Huang, N.~Ahuja, and M.-H. Yang, ``Deep laplacian pyramid
  networks for fast and accurate super-resolution,'' in \emph{Proceedings of
  the IEEE conference on computer vision and pattern recognition}, 2017, pp.
  624--632.

\bibitem{SRGAN}
C.~Ledig, L.~Theis, F.~Husz{\'a}r, J.~Caballero, A.~Cunningham, A.~Acosta,
  A.~Aitken, A.~Tejani, J.~Totz, Z.~Wang \emph{et~al.}, ``Photo-realistic
  single image super-resolution using a generative adversarial network,'' in
  \emph{Proceedings of the IEEE conference on computer vision and pattern
  recognition}, 2017, pp. 4681--4690.

\bibitem{RDN}
Y.~Zhang, Y.~Tian, Y.~Kong, B.~Zhong, and Y.~Fu, ``Residual dense network for
  image super-resolution,'' in \emph{CVPR}, 2018.

\bibitem{FSRCNN}
C.~Dong, C.~C. Loy, and X.~Tang, ``Accelerating the super-resolution
  convolutional neural network,'' in \emph{European conference on computer
  vision}.\hskip 1em plus 0.5em minus 0.4em\relax Springer, 2016, pp. 391--407.

\end{thebibliography}

\end{document}